\documentclass[lettersize,journal]{IEEEtran}
\usepackage{amsmath,amsfonts}

\usepackage{algorithm}
\usepackage{array}
\usepackage[caption=false,font=normalsize,labelfont=sf,textfont=sf]{subfig}
\usepackage{textcomp}
\usepackage{stfloats}
\usepackage{url}
\usepackage{verbatim}
\usepackage{graphicx}
\usepackage{cite}
\usepackage{times}
\usepackage{epsfig}
\usepackage{amssymb}

\usepackage{booktabs}
\usepackage{multirow}
\usepackage{alltt}
\usepackage{algpseudocode}
\usepackage{algorithm}
\usepackage{color}
\usepackage[dvipsnames]{xcolor,colortbl}
\usepackage{pgfplots}
\usepackage{caption}
\usepackage{subcaption}

\makeatletter
\def\ps@IEEEtitlepagestyle{%
  \def\@oddfoot{\mycopyrightnotice}%
  \def\@oddhead{\hbox{}\@IEEEheaderstyle\leftmark\hfil\thepage}\relax
  \def\@evenhead{\@IEEEheaderstyle\thepage\hfil\leftmark\hbox{}}\relax
  \def\@evenfoot{}%
}
\def\mycopyrightnotice{%
  \begin{minipage}{\textwidth}
  \centering \scriptsize
  IEEE Copyright Notice~\copyright~2024 IEEE. Personal use of this material is permitted.  Permission from IEEE must be obtained for all other uses, in any current or future media, including reprinting/republishing this material for advertising or promotional purposes, creating new collective works, for resale or redistribution to servers or lists, or reuse of any copyrighted component of this work in other works.
  \end{minipage}
}
\makeatother

\begin{document}

\title{Robustness-Reinforced Knowledge Distillation \\ with Correlation Distance and Network Pruning}

\author{Seonghak Kim, Gyeongdo Ham, Yucheol Cho, and Daeshik Kim\\
\thanks{Seonghak Kim, Gyeongdo Ham, Yucheol Cho and Daeshik Kim are with the School of Electrical Engineering, Korea Advanced Institute of Science and Technology (KAIST), Daejeon 34141, Republic of Korea. \protect\\
E-mail: \{hakk35, rudeh6185, yc\_cho, daeshik\}@kaist.ac.kr}
\thanks{Seonghak Kim, Gyeongdo Ham, and Yucheol Cho contributed equally to this work. Daeshik Kim is a corresponding author.}
\thanks{This is the author’s version of an article that has been accepted for publication in IEEE Transactions on Knowledge and Data Engineering.}
\thanks{Digital Object Identifier (DOI): 10.1109/TKDE.2024.3438074}
}

\markboth{IEEE Transactions on Knowledge and Data Engineering}%
{Shell \MakeLowercase{\textit{et al.}}: A Sample Article Using IEEEtran.cls for IEEE Journals}


\maketitle

\begin{abstract}
The improvement in the performance of efficient and lightweight models (i.e., the student model) is achieved through knowledge distillation (KD), which involves transferring knowledge from more complex models (i.e., the teacher model). However, most existing KD techniques rely on Kullback-Leibler (KL) divergence, which has certain limitations. First, if the teacher distribution has high entropy, the KL divergence's mode-averaging nature hinders the transfer of sufficient target information. Second, when the teacher distribution has low entropy, the KL divergence tends to excessively focus on specific modes, which fails to convey an abundant amount of valuable knowledge to the student. Consequently, when dealing with datasets that contain numerous confounding or challenging samples, student models may struggle to acquire sufficient knowledge, resulting in subpar performance. Furthermore, in previous KD approaches, we observed that data augmentation, a technique aimed at enhancing a model's generalization, can have an adverse impact. Therefore, we propose a Robustness-Reinforced Knowledge Distillation (R2KD) that leverages correlation distance and network pruning. This approach enables KD to effectively incorporate data augmentation for performance improvement. Extensive experiments on various datasets, including CIFAR-100, FGVR, TinyImagenet, and ImageNet, demonstrate our method's superiority over current state-of-the-art methods.

\end{abstract}


\section{Introduction}
\IEEEPARstart{T}{he} remarkable progress of computer vision in recent years, powered by deep neural networks, has enabled better performance in practical applications such as classification, object detection, and semantic segmentation. However, to ensure the effective functionality of these vision tasks on mobile or low-capacity devices, it is important to consider the limited computational resources available. Various model compression techniques, including model quantization, model pruning, and knowledge distillation, have emerged as crucial research areas to address this challenge. Among these techniques, Knowledge distillation (KD) facilitates smaller networks, known as a student model, to have comparable performance to larger networks, known as a teacher model. This is accomplished by transferring knowledge from a teacher model to a student model, which can be used practically in place of the larger network~\cite{application1, application2, application3}. In KD, the term “\textit{knowledge}” refers to intermediate feature maps, class predictions as soft labels, or penultimate layer representations.

{Hinton’s foundational paper on knowledge distillation~\cite{hinton} introduced the idea of softening probabilities by using a higher-than-normal temperature (e.g., $T=4$) in the softmax function. The generalization tendencies of complicated model can be better understood by considering the relative probabilities of non-target classes. Since then, \textit{dark knowledge} has been used to describe this information in soft targets~\cite{dark1, dark2,dark3}. Building on this research, we have also embraced the term ‘dark knowledge’ to describe the helpful information transferred from the teacher to the student model. }

Even after the introduction of Vanilla KD, most logit-based KDs still rely on KL divergence to measure the similarity between the soft probabilities generated by the teacher and student models. However, KL divergence-based KDs inherently harbor potential drawbacks that can impede improvements in the student model's performance. {We have conceptually identified that KL divergence can lead the student's predictions to either \textit{mode averaging} (as is commonly well-known~\cite{wen2023f}) or \textit{mode focusing} (as is indirectly expressed in the formula of~\cite{dkd}), depending on the entropy of the teacher's predictions.}

\begin{figure}[t!]
  \centering
  \includegraphics[width=0.5\textwidth]{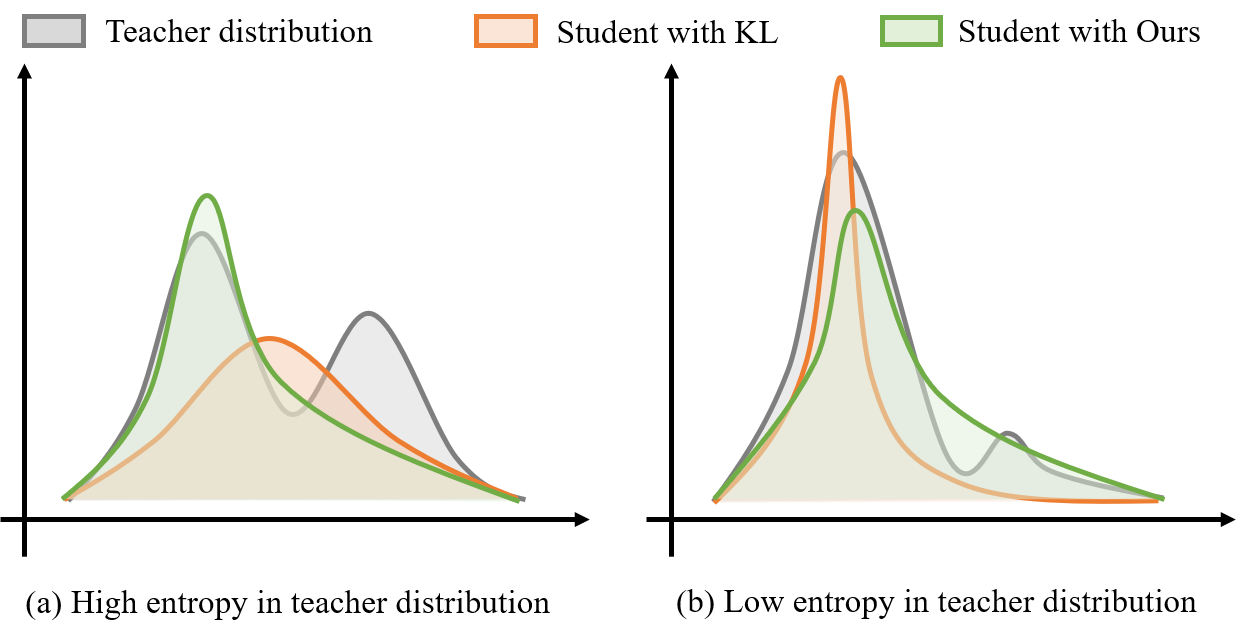}
  \caption{\textbf{Comparison of distribution} with (a) high entropy in teacher distribution and (b) low entropy in teacher distribution. Gray: teacher distribution. Orange: student distribution with KL divergence. Green: student distribution with our correlation distance.}
  \label{fig:fkl}
\end{figure}


{To overcome the limitations of KL divergence, we propose a method for students to learn independently of the teacher model's prediction entropy. This method involves projecting the predictions of both the teacher and the student into a vector space of the same dimension, aiming to align the student vector with the teacher vector. To measure vector similarity, we utilize the commonly used metric, value-based correlations. Furthermore, we enhance performance by incorporating rank-based correlations to reflect non-linear relationships between student and teacher vectors.}

{As shown in Fig.~\ref{fig:fkl}, when the teacher model has high entropy (indicating low confidence), our method encourages the student to focus on acquiring target information, while KL divergence causes the student model to receive insufficient critical target information due to its mode-averaging property, simultaneously acquiring unnecessary non-target information. Conversely, when the entropy is low (indicating high confidence), our method promotes the student's focus on learning dark knowledge, whereas KL divergence causes the student model to receive an inadequate amount of dark knowledge due to its mode-focusing property. Consequently, while KL divergence leads the student model to obtain inappropriate knowledge from the teacher and negatively affects its performance, our method addresses this by using the prediction in vector space. We will explain the limitations of KL divergence in more detail in Sec.~\ref{subsec:limitation} and discuss the differences in entropy among KD methods in Sec.~\ref{sec:experiment}}

\begin{figure*}[]
  \centering
  \includegraphics[width=1.0\textwidth]{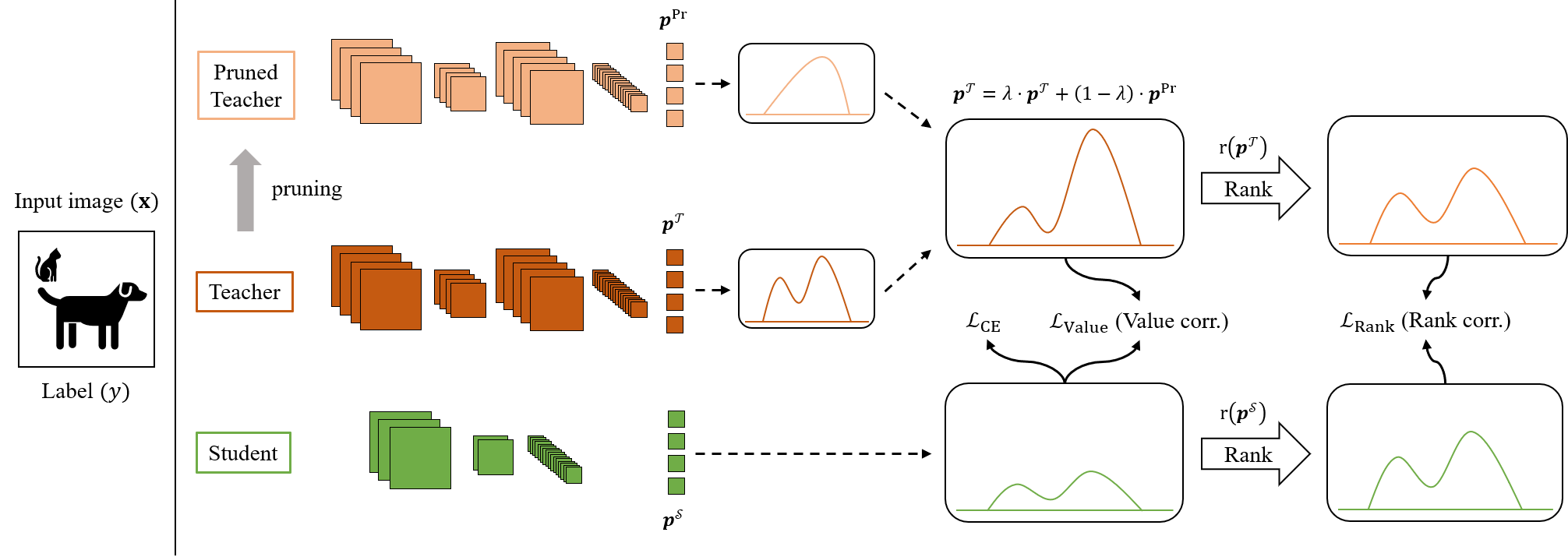}
  \caption{\textbf{Illustration of the proposed method.} The pruned teacher model is a duplicate of the pre-trained teacher model, and the input image is passed through these two models to produce $p^{\mathrm{Pr}}$ and $p^{\mathcal{T}}$ predictions. To address uncertainty in images, these two predictions are combined into a single prediction $p^{\mathcal{T}}$, which is then used to distill knowledge into the student model. We consider both $p^{\mathcal{S}}$ and $p^{\mathcal{T}}$ as vectors, and employ value- and rank-based correlation techniques to make $p^{\mathcal{S}}$ resemble $p^{\mathcal{T}}$.}
  \label{fig:overview}
\end{figure*}

{Furthermore, we apply network pruning to the teacher model to enhance the robustness of the student model when dealing with challenging and heavily augmented images. Integrating the pruned model allows the teacher to eliminate superfluous knowledge that is difficult for the student to grasp, ensuring the conveyance of only valuable information. While network pruning is typically used for model compression, in this paper, we employ it to identify challenging images. Our motivation stems from the fact that pruning removes specific weights from the original model, thus increasing the discrepancy in probability distribution compared to the original model when handling ambiguous images. We design a modified teacher prediction by taking the weighted sum of the predictions from both the pruned teacher model and the original teacher model. This combined prediction is then utilized for vector alignment with the student's prediction. Our approach ensures that the teacher imparts valuable knowledge to the student, even when dealing with easily distinguishable images, as the predictions from the pruned and original teacher models closely align, resulting in a highly confident target prediction. Notably, using the pruned model does not require any additional training process since it utilizes the existing pre-trained teacher model as-is.}



To demonstrate the effectiveness and robustness of our method in handling challenging and heavily augmented images, we applied {several data augmentation methods~\cite{cutout,mixup,cutmix}} to various datasets, including CIFAR100, FGVR, TinyImageNet, and ImageNet. Through extensive experiments, our approach outperformed other methods, not only on standard datasets but also on augmented datasets. Unlike other methods where the use of data augmentation, designed to enhance generalization and consequently improve the model's performance, had negative impacts on student performance, our approach consistently enhances performance without such constraints. As a result, as shown in Fig.~\ref{fig:overview}, our innovative KD approach, integrating value-based correlations, rank-based correlations, and network pruning, effectively improves student accuracy and robustness, providing a solid foundation for integrating data augmentation into knowledge distillation.

Our contributions can be highlighted as follows:
\begin{itemize}
    \item We show that {high} teacher's entropy leads to insufficient target information for the student, while {low} entropy results in inadequate dark knowledge transfer, both negatively impacting the student's performance.

    \item We propose a novel methodology using correlation distance to capture both linear and non-linear relationships between teacher and student models, improving knowledge distillation.
    
    \item We apply network pruning to the teacher model to enhance the student model's robustness, particularly with challenging images, without requiring additional training.
    
    \item Through extensive experiments, we demonstrate the methodology's effectiveness, even with challenging and heavily augmented images, making it a valuable approach for integrating data augmentation into knowledge distillation.
    
\end{itemize}

\section{Related Works}

\subsection{Knowledge Distillation}


Knowledge distillation (KD) offers a solution by transferring knowledge from a more complex and high-performing network to a smaller, more efficient network. Over the years, there has been a surge of research in KD and the development of better distillation techniques. {Since the concept of KD was first introduced by Hinton~\cite{hinton}, it has expanded into three major approaches: logits-based~\cite{hinton, dml, takd, dkd}, feature-based~\cite{fitnet, PKT, CRD, AT, VID, heo2019comprehensive, reviewkd}, and relation-based~\cite{RKD, masckd, ckd_mkt} distillation. Conventional logits- and feature-based KDs independently convey the knowledge extracted from each data sample. In contrast, relation-based KDs extract and transfer the relational knowledge between multiple samples. For example, MASCKD~\cite{masckd} utilizes attention maps from multiple intermediate layers to form sample correlations, effectively addressing the important sample regions that previous relational KD methods fail to capture. More recently, CKD-MKT~\cite{ckd_mkt} integrates knowledge from both individual instances and instance relations, overcoming the limitation of existing methods that typically consider only one type of knowledge from either instance features or relations.}

{However, because our method considers the model’s prediction obtained from each sample, we focus on individual instance-based distillation rather than instance relation-based distillation.} While feature-based distillation allows students to learn a wider range of information compared to logit-based distillation, it has limited practical applicability due to challenges related to accessing the intermediate layer in real-world scenarios, primarily because of privacy and security concerns~\cite{jin2023multi}. Therefore, our focus is on logit-based distillation, which is more suitable for practical use.

The majority of logit-based distillation methods employ the Kullback-Leibler (KL) divergence to align the probability distributions between teacher and student models, representing the simplest and most straightforward approach to knowledge transfer in KD. However, depending on the entropy of the teacher's distribution, students using KD are prone to receiving unintended information from the teacher's distribution. In this paper, we conceptually describe the potential student's distribution based on teacher entropy and utilize a correlation-based distance to overcome this issue.

\subsection{Correlation-based Distance}
The correlation is a commonly employed technique in clustering, used to distinguish groups with similar data characteristics and assign them to distinct clusters~\cite{correlation_book, correlation_paper}. There are two types of correlation: value-based correlation (e.g., Eisen and Pearson~\cite{pearson}) and rank-based correlation (e.g., Spearman~\cite{spearman} and Kendall~\cite{kendall}). A perfect correlation between two random variables yields a correlation coefficient of 1, whereas no correlation between them results in a coefficient of 0.

In the context of knowledge distillation, it is well-established that the performance of the student model depends on receiving appropriate target information and dark knowledge from the teacher model~\cite{ATS}. This is more critical than solely having highly confident target predictions (i.e., achieved through low-temperature scaling) or excessively high dark knowledge (i.e., through high-temperature scaling) from the teacher model.

While utilizing only value-based correlation provides valuable linear information between teacher and student distributions, it has a limitation in that it cannot capture nonlinear relationships. This implies that it does not facilitate the optimal transfer of target information and dark knowledge from the teacher to the student. Therefore, incorporating both linear and non-linear correlations between the teacher and student distributions can help the student acquire the optimal target information and dark knowledge.

\subsection{Network Pruning}

Network pruning involves eliminating unnecessary weights while preserving crucial ones to compress a model without compromising its accuracy. Traditionally, network pruning was mainly employed in scenarios with limited computational resources. However, recent studies have employed network pruning for a different purpose: identifying and filtering hard-to-memorize samples. In their work,~\cite{hooker} introduced the concept of Pruning Identified Exemplars (PIEs) and demonstrated that PIEs exhibit distinct characteristics, such as corrupted images, fine-grained classification, and abstract representations. Leveraging these characteristics,~\cite{selfdamaging,artkd} utilized a dynamic self-competitive model to detect confusing samples, opposing the original target model.
Additionally,~\cite{ppesudo} highlighted an issue related to biased models in the easy class when assigning pseudo-labels based solely on a single model's confidence scores during pseudo-labeling. To address this problem, they introduced the concept of an Easy-to-Forget (ETF) sample finder and explained how to incorporate it into the learning process. Building on the insights from these studies, our method employs soft label distillation by combining pruned and original teacher outputs, resulting in a more robust framework, even when addressing challenging and highly augmented samples. Numerous experimental results, including standard benchmark datasets and augmented datasets~\cite{empirical, relatedda, relatedda2, relatedda3,cutmix,mixup}, demonstrate that our approach outperforms other knowledge distillation methods (KD).

\section{Proposed Method}
\label{sec:method}

\subsection{Limitation of KL divergence}
\label{subsec:limitation}

The majority of current logits-based KDs utilize the KL divergence to instruct the student model in capturing the teacher model's distribution. This can be represented as follows:

\begin{equation}
\mathcal{D}_{\mathrm{KL}}\left(p^{\mathcal{T}} \| p^{\mathcal{S}}\right)=\sum_{i=1}^C p_i^{\mathcal{T}} \log \left(\frac{p_i^{\mathcal{T}}}{p_i^{\mathcal{S}}}\right)
\end{equation}
where $p_i^{\mathcal{S}}$ and $p_i^{\mathcal{T}}$ represent the probability associated with the $i-$th class of the teacher and student model, respectively and $C$ denote the class number.

In an optimization process, as the KL divergence decreases, the student's ability to mimic the teacher distribution improves. Nevertheless, given that the student model inherently possesses a limited capacity to represent the distribution compared to teacher model, the manner in which the student approximates the distribution will differ based on the entropy of the teacher's distribution. To grasp this concept, we can examine two scenarios for $p_i^{\mathcal{T}}$, specifically, when $p_i^{\mathcal{T}}$ equals $0$ and when $p_i^{\mathcal{T}}$ is greater than $0$, in order to predict how $p_i^{\mathcal{S}}$ tends to be approximated.

\subsubsection{Case I, $p_i^{\mathcal{T}}=0$}
Because $p_i^{\mathcal{T}}$ represents the weight of the difference between $p_i^{\mathcal{T}}$ and $p_i^{\mathcal{S}}$, the loss consistently remains at its minimum value, regardless of the difference between the values of $p_i^{\mathcal{T}}$ and $p_i^{\mathcal{S}}$. In other words, when $p_i^{\mathcal{T}}$ equals 0, it has no impact on the loss value, no matter how much $p_i^{\mathcal{S}}$ deviates from $p_i^{\mathcal{T}}$.

\subsubsection{Case II, $p_i^{\mathcal{T}}>0$}
On the flip side, in this case, the value of term $\log \left(\frac{p_i^{\mathcal{T}}}{p_i^{\mathcal{S}}}\right)$ will have an impact on the loss. In other words, when $p_i^{\mathcal{T}}$ is greater than 0, it is advisable to minimize the difference between $p_i^{\mathcal{T}}$ and $p_i^{\mathcal{S}}$ to reduce the loss as much as possible.

For this reason, KL divergence exhibits a mode-averaging property, as depicted in Fig.~\ref{fig:fkl}(a). However, this behavior depends on the teacher's entropy. When the teacher has high entropy (Fig.~\ref{fig:fkl}(a)), concentrating on a single mode of the teacher increases the difference between the other modes and the student's mode, justifying mode averaging. Conversely, when the teacher's entropy is low (Fig.~\ref{fig:fkl}(b)), focusing on one mode may result in a smaller overall loss, as the difference between the other modes and the student's mode is smaller.

\begin{figure*}[]
  \centering
  \includegraphics[width=1.0\textwidth]{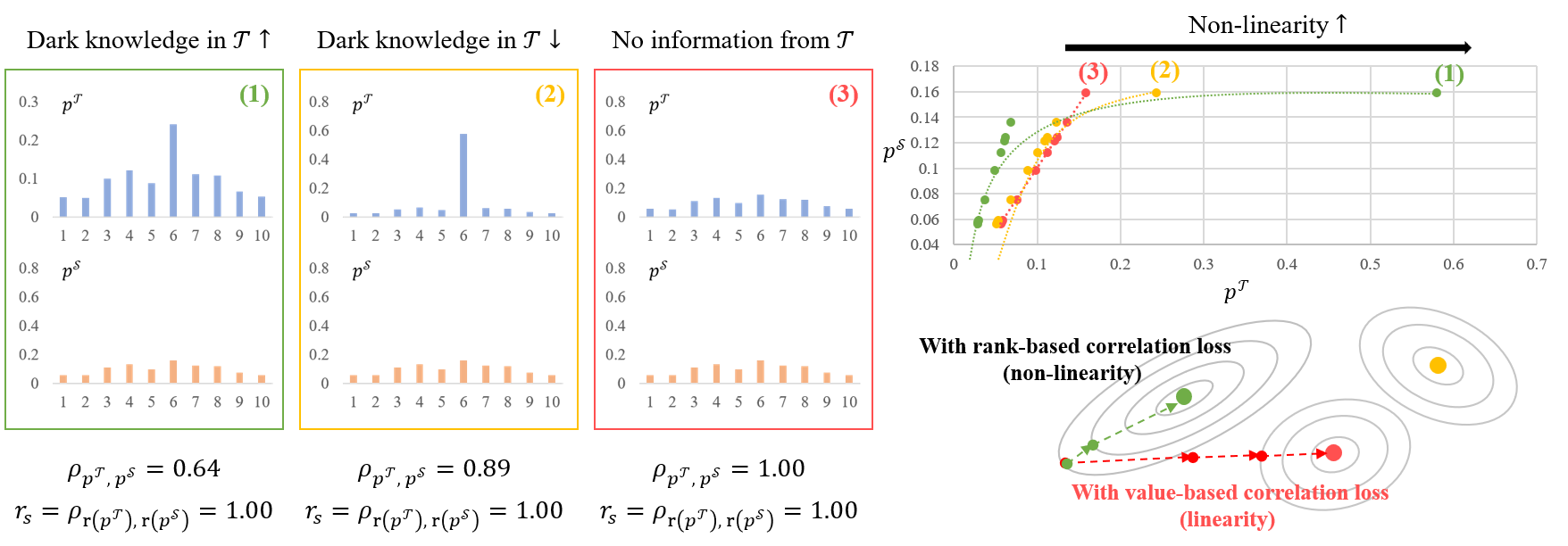}
  \caption{\textbf{Understanding of correlation coefficient.} Value-based correlation coefficient (denoted as $\rho_{p^{\mathcal{T}}, p^{\mathcal{S}}}$) and rank-based correlation coefficient (denoted as $r_s$) between teacher and student predictions. When only a value-based correlation is applied in KDs, the student's weights are updated to be completely matched with the teacher's predictions (red line, marked as (3)). However, when a rank-based correlation is also applied, the student model can learn to obtain rich information from the teacher (green line, marked as (1)).}
  \label{fig:correlation}
\end{figure*}

However, in the context of distillation, these properties of the KL divergence can result in unfavorable outcomes for the student model. When the teacher's entropy is high, there is a need to distill more information about the target prediction due to the scarcity of target-related information. Conversely, when the teacher's entropy is low, it makes sense to convey a surplus of dark knowledge related to non-target classes, given that target information is already abundant. As a solution to the challenges posed by the KL divergence, we treat the teacher's and student's distributions as vectors and aim to align the direction of the student's vector with that of the teacher.

\subsection{Correlation Distance Loss}
\label{subsec:dist}

Our objective is to utilize value-based and rank-based correlation distance to ensure an optimal alignment between the student and teacher distributions. Value-based correlation and Rank-based correlation can be explained using cosine similarity, {$\operatorname{Sim}\left(\cdot \right)$,} as follows:

\begin{equation}
\operatorname{Sim}\left({\boldsymbol{p}^{\mathcal{T}}, \boldsymbol{p}^{\mathcal{S}}}\right)=\frac{\sum_{i=1}^n p_i^{\mathcal{T}} p_i^{\mathcal{S}}}{\sqrt{\sum_{i=1}^n\left(p_i^{\mathcal{T}}\right)^2 \sum_{i=1}^n\left(p_i^{\mathcal{S}}\right)^2}}
\end{equation}

\begin{equation}
\rho_{\boldsymbol{p}^{\mathcal{T}}, \boldsymbol{p}^{\mathcal{S}}}=\operatorname{Sim}\left(\boldsymbol{p}^{\mathcal{T}}-\overline{\boldsymbol{p}^{\mathcal{T}}}, \boldsymbol{p}^{\mathcal{S}}-\overline{\boldsymbol{p}^{\mathcal{S}}}\right)
\end{equation}

\begin{equation}
r_s=\rho_{\mathrm{r}\left(\boldsymbol{p}^{\mathcal{T}}\right), \mathrm{r}\left(\boldsymbol{p}^{\mathcal{S}}\right)}
\end{equation}
{Here, $\rho_{\boldsymbol{p}^{\mathcal{T}}, \boldsymbol{p}^{\mathcal{S}}}$ represents the value-based correlation between the predictions from the teacher  $\left({\boldsymbol{p}}^\mathcal{T}\in \mathbb{R}^C\right )$ and student $\left({\boldsymbol{p}}^\mathcal{S}\in \mathbb{R}^C \right)$ models for a single sample. $\overline{\boldsymbol{p}^{\mathcal{T}}}$ and $\overline{\boldsymbol{p}^{\mathcal{S}}}$ denotes the prediction average $\left(\overline{\boldsymbol{p}}\in \mathbb{R}  \right)$. $r_s$ is the rank-based correlation, with $\mathrm{r}(\boldsymbol{p})$ meaning $\mathrm{rank}\left({\boldsymbol{p}}\right)$}.


In order to increase the loss for weaker correlation, we utilize the correlation distance as follows:

\begin{equation}
    d_{\mathrm{Value}}(\boldsymbol{p}^\mathcal{T}, \boldsymbol{p}^\mathcal{S}) = 1 - \operatorname{Sim}\left({\boldsymbol{p}^{\mathcal{T}}, \boldsymbol{p}^{\mathcal{S}}}\right)
\end{equation}

\begin{equation}
    d_{\mathrm{Rank}}(\boldsymbol{p}^\mathcal{T}, \boldsymbol{p}^\mathcal{S}) = 1 - r_s.
\end{equation}

Although some previous research utilizes the Pearson distance~\cite{cao2022pkd}, one of the widely adopted value-based correlation distances, it proves inadequate in capturing non-linear relationships between teacher and student predictions due to its intrinsic linearity, and it is susceptible to outlier values~\cite{kim2015instability}.

Fig.~\ref{fig:correlation} illustrates the distinction between linear and non-linear relationships in two probability distributions. We distinguish between the probability distributions of the student and teacher models in three scenarios: (1) when the teacher possesses optimal dark knowledge for distillation, (2) when the teacher has a high confidence score but low dark knowledge, and (3) when the teacher's information minimally impacts the student model's performance. Subsequently, we compute both the value-based correlation (denoted as $\rho_{\boldsymbol{p}^{\mathcal{T}}, \boldsymbol{p}^{\mathcal{S}}}$) and rank-based correlation (denoted as $r_s$) for each case.

Value-based correlation produces different values across these cases, as shown in the graph on the right. In contrast, rank-based correlation maintains a consistent value. Notably, value-based correlation equals one only when student and teacher model outcomes are linear (i.e., correlation distance becomes minimum), whereas rank-based correlation can reach one when outcomes exhibit a monotonically increasing trend, even if they aren't linear. Consequently, relying solely on value-based correlation as a loss function would lead to the student model's weight being updated in the red route, even though the green route contains the most valuable knowledge for distillation. This results in inadequate knowledge transfer to the student model, ultimately leading to its poor performance. To address this issue, we introduce rank-based correlation as a new loss, allowing us to prioritize the process in the green route rather than the red route.

For our method, \textit{Robustness-Reinforced Knowledge Distillation (R2KD)}, the final objective function is designed as follows:

\begin{equation}
\mathcal{L}_{\text{R2KD}}= \mathcal{L}_{\text{CE}}+\alpha \mathcal{L}_{\text{Value}}+\beta \mathcal{L}_{\text{Rank}}
\end{equation}

\begin{equation}
\mathcal{L}_{\text{Value}}=\frac{1}{B} \sum_{i=1}^B d_{\mathrm{Value}}\left(\boldsymbol{p}^{\mathcal{T}}, \boldsymbol{p}^{\mathcal{S}}\right)\end{equation}

\begin{equation}
\mathcal{L}_{\text{Rank}}=\frac{1}{B} \sum_{i=1}^B d_{\mathrm{Rank}}\left(\boldsymbol{p}^{\mathcal{T}}, \boldsymbol{p}^{\mathcal{S}}\right).
\end{equation}
{where $\mathcal{L}_{\text{CE}}$ is cross-entropy loss; $\alpha$ and $\beta$ are hyper-parameters representing the weights of the value-based and rank-based losses, respectively.}

We demonstrate that our method exhibits robust performance not only on standard datasets (e.g., CIFAR-100) but also on datasets containing challenging and confusing samples (e.g., TinyImageNet, ImageNet and FGVR), even in scenarios involving data augmentation. This is evident in Sec.~\ref{sec:experiment}. To further enhance the robustness of our model, we also incorporate a pruned teacher model, as elaborated in the following Sec.~\ref{subsect:pruned}.

\subsection{Pruned Teacher Network}
\label{subsect:pruned}
According to Hooker's findings~\cite{hooker}, the pruned teacher model has the property of losing its ability to remember difficult-to-retain samples. Hence, we can obtain refined teacher predictions, which reduces the confidence of predictions for challenging samples while retaining the confidence of easy samples. The predictions of teacher $\boldsymbol{p}^\mathcal{T}$ in our loss function can be achieved as follows:

\begin{equation}
\boldsymbol{p}^\mathcal{T} = \lambda \cdot \boldsymbol{p}^\mathcal{T} + (1-\lambda) \cdot \boldsymbol{p}^\text{Pr}, 
\end{equation}

where weighting value $\lambda$ ($0< \lambda <1$) is a hyper-parameter and $\boldsymbol{p}^\text{Pr}$ means predictions from the pruned teacher model, the $\lambda$ for all experiments are shown in the supplemental material.
The purpose of this ensemble that utilizes the knowledge of the original and the pruned teacher is distinct from the general ensemble method, which seeks to utilize the knowledge of multiple models with different information. Although the pruned model typically exhibits inferior performance to the non-pruned model without retraining, we can take advantage of the fact that the pruned model's predictions on challenging samples follow a different distribution from the non-pruned model's predictions. Consequently, by combining the two predictions, we can maintain high confidence scores for simple samples while reducing confidence scores for difficult samples. We apply this combined knowledge to our objective function for distillation. As a result, our approach can mitigate the risk of direct distillation to student models in situations where the teacher model's predictions are incorrect for challenging samples. These properties are even more effective in knowledge distillation with data augmentation. 

\section{Experiment}
\label{sec:experiment}

We assess the effectiveness of our method by comparing it with other knowledge distillation approaches, including both logit- and feature-based methods, across a variety of architectural networks and image classification datasets. Furthermore, we employ data augmentation techniques for each dataset, thereby demonstrating the superior robustness of our method when compared to others.

\begin{table*}[]
\caption{\textbf{Results on classification.} Top-1 accuracy (\%) on the CIFAR-100 testsets when using teacher and student models with the same architectures. The best results are highlighted in \textbf{bold} and the second best \underline{underlined}. The number in parentheses indicates the standard deviation and $\Delta$ represents the performance difference between the best result of previous KDs and our best result. The results marked as * was not in their paper, so we conducted three new runs and then calculated the average.}
\centering
\resizebox{1.0\textwidth}{!}{%
    \begin{tabular}{@{}ccccccccccc@{}}
        \toprule
        \multirow{4}{*}{Distillation} & \multirow{2}{*}{Teacher}&\multicolumn{1}{c}{WRN-40-2} & \multicolumn{1}{c}{WRN-40-2}  & \multicolumn{1}{c}{ResNet56}  &   \multicolumn{1}{c}{ResNet110}    &\multicolumn{1}{c}{ResNet32x4}  &   \multicolumn{1}{c}{VGG13} &  \multirow{4}{*}{Avg.}\\ 
         &&\multicolumn{1}{c}{75.61} & \multicolumn{1}{c}{75.61}    &\multicolumn{1}{c}{72.34}  &   \multicolumn{1}{c}{74.31}  &  \multicolumn{1}{c}{79.42}  &   \multicolumn{1}{c}{74.64}  \\
        
        &\multirow{2}{*}{Student}& \multicolumn{1}{c}{WRN-16-2} & \multicolumn{1}{c}{WRN-40-1}  &  \multicolumn{1}{c}{ResNet20}  &   \multicolumn{1}{c}{ResNet32}  &  \multicolumn{1}{c}{ResNet8x4}  &   \multicolumn{1}{c}{VGG8}  \\ 
        &&  \multicolumn{1}{c}{73.26} & \multicolumn{1}{c}{71.98}  &  \multicolumn{1}{c}{69.06}  &   \multicolumn{1}{c}{71.14}  &  \multicolumn{1}{c}{72.50}  &   \multicolumn{1}{c}{70.36} \\
        \midrule
        \multirow{9}{*}{Features} & FitNet~\cite{fitnet}                
        & 73.58 & 72.24 & 69.21 & 71.06 & 73.50 & 71.02 & 71.77 \\
        & PKT~\cite{PKT}                     
        &  74.54 & 73.54 & 70.34 & 72.61 & 73.64 & 72.88 & 72.93 \\
        & RKD~\cite{RKD}                     
        &  73.35 & 72.22 & 69.61 & 71.82 &  71.90 & 71.48 & 
        71.73 \\
        & CRD~\cite{CRD}                     
        &  75.48 & 74.14 & 71.16 & 73.48 & 75.51 & 73.94  &
        73.95 \\   
        & AT~\cite{AT}                     
        &  74.08 & 72.77 & 70.55 & 72.31 &  73.44 & 71.43  &
        72.43 \\
        & VID~\cite{VID}                     
        &  74.11 & 73.30 & 70.38 & 72.61 &  73.09 & 71.23  &
        72.45 \\
        & OFD~\cite{heo2019comprehensive}                    
        &  75.24 & 74.33 & 70.98 & 73.23 &  74.95 & 73.95  &
        73.78 \\
        & SP~\cite{sp}                    
        &  73.83 & 72.43 & 69.67 & 72.69 & 72.94 & 72.68  & 72.37 \\        
        & ReviewKD~\cite{reviewkd}            
        &  76.12 & 75.09 & 71.89 & 73.89 &  75.63 & 74.84  &
        74.58 \\
        \midrule
        \multirow{6}{*}{Logits} & KD~\cite{hinton}                      
        & 74.92 & 73.54 & 70.66 & 73.08 & 73.33 & 72.98 & 73.06\\
& DML~\cite{dml}                      
        & 73.58 & 72.68 & 69.52 & 72.03 & 72.12 & 71.79 & 71.95\\
& TAKD~\cite{takd}                      
        & 75.12 & 73.78 & 70.83 & 73.37 & 73.81 & 73.23 & 73.36\\
        & DKD~\cite{dkd}               
        &  76.24 & 74.81 & 71.97 & 74.11 &  76.32 & 74.68 & 74.69 \\ 

        & {SimKD~\cite{simkd}}                    
        &  {75.96*} & {75.18*} & {68.71*} & {72.17*} &   {\textbf{78.08}} & {74.93}  &
        {74.17} \\
        
        & {DIST~\cite{dist}}   
        &  {-} & {74.73} & {71.75} & {-} &  {76.31} & {-} & {74.26} \\ 
        
        \midrule
        &\textbf{R2KD}   
        &  {76.62 (0.11)} & {75.24 (0.06)} & {72.42 (0.04)} & {74.09 (0.10)} & {77.01 (0.16)} & {75.26 (0.04)} & 75.10 \\
        &\textbf{{R2KD w/ MixUp}}   
        &  {\underline{76.94} (0.16)} & {\underline{76.02} (0.08)} & {\underline{72.60} (0.06)} & {\underline{74.66} (0.08)} & { {\underline{78.07}} (0.12)} & {\underline{75.65} (0.08)} & {\underline{75.66}} \\ 
        &\textbf{R2KD w/ CutMix}   
        &  {\textbf{77.06} (0.11)} & {\textbf{76.21} (0.11)} & {\textbf{72.65} (0.09)} & {\textbf{75.04} (0.12)} & { {77.70} (0.15)} & {\textbf{76.40} (0.16)} & \textbf{75.84} \\ 
        &$\Delta$  
        &  \textbf{+0.82} & {\textbf{+1.03}} & \textbf{+0.68} & \textbf{+0.93} & {{-0.01}} & {\textbf{+1.47}} & \textbf{+1.15}\\
        \bottomrule
    \end{tabular}
}
\label{tab:cifar_homogeneous}
\end{table*}

\begin{table*}[]
\caption{\textbf{Results on classification.} Top-1 accuracy (\%) on the CIFAR-100 testsets when using teacher and student models with different architectures. The best results are highlighted in \textbf{bold} and the second best \underline{underlined}. {The number in parentheses indicates the standard deviation and $\Delta$ represents the performance difference between the best result of previous KDs and our best result. The results marked as * was not in their paper, so we conducted three new runs and then calculated the average.}}
\centering
\resizebox{1.0\textwidth}{!}{%
    \begin{tabular}{@{}cccccccccc@{}}
    \toprule
        \multirow{4}{*}{Distillation} & \multirow{2}{*}{Teacher}&  \multicolumn{1}{c}{WRN-40-2} & \multicolumn{1}{c}{ResNet50}  & \multicolumn{1}{c}{ResNet32x4}  &   \multicolumn{1}{c}{ResNet32x4}    &\multicolumn{1}{c}{VGG13} & \multirow{4}{*}{Avg.} \\ 
         && \multicolumn{1}{c}{75.61} & \multicolumn{1}{c}{79.34}    &\multicolumn{1}{c}{79.42}  &   \multicolumn{1}{c}{79.42}  &  \multicolumn{1}{c}{74.64}  \\
        
        &\multirow{2}{*}{Student}& \multicolumn{1}{c}{ShuffleNet-V1} & \multicolumn{1}{c}{MobileNet-V2}  &  \multicolumn{1}{c}{ShuffleNet-V1}  &   \multicolumn{1}{c}{ShuffleNet-V2}  &  \multicolumn{1}{c}{MobileNet-V2} \\ 
        &&  \multicolumn{1}{c}{70.50} & \multicolumn{1}{c}{64.60}  &  \multicolumn{1}{c}{70.50}  &   \multicolumn{1}{c}{71.82}  &  \multicolumn{1}{c}{64.60}  \\
        \midrule
        \multirow{9}{*}{Features} & FitNet~\cite{fitnet}                
        & 73.73 & 63.16 & 73.59 & 73.54 & 64.14 & 69.63\\
        &PKT~\cite{PKT}                     
        &  73.89 & 66.52 & 74.10 & 74.69 & 67.13  & 71.27 \\
        &RKD~\cite{RKD}                     
        &  72.21 & 64.43 & 72.28 & 73.21 &  64.52 & 69.33 \\
        &CRD~\cite{CRD}                     
        &  76.05 & 69.11 & 75.11 & 75.65 & 69.73  & 73.13 \\        
        &AT~\cite{AT}                     
        &  73.32 & 58.58 & 71.73 & 72.73 &  59.40  & 67.15 \\
        &VID~\cite{VID}                     
        &  73.61 & 67.57 & 73.38 & 73.40 &  65.56  & 70.70 \\
        &OFD~\cite{heo2019comprehensive}                     
        &  75.85 & 69.04 & 75.98 & 76.82 &  69.48  & 73.43 \\
        &{SP~\cite{sp}}                     
        &  {74.52} & {68.08} & {73.48} & {74.56} &  {66.30}  & {71.39} \\
        
        &ReviewKD~\cite{reviewkd}                      
        &  77.14 & 69.89 & 77.45 & 77.78 &  70.37 & 74.53  \\
        \midrule
        \multirow{6}{*}{Logits} & KD~\cite{hinton}                      
        & 74.83 & 67.35 & 74.07 & 74.45 & 67.37 & 71.60 \\
        & DML~\cite{dml}               
        &  72.76 & 65.71 & 72.89 & 73.45 &  65.63 & 70.09 \\
        & TAKD~\cite{takd}               
        &  75.34 & 68.02 & 74.53 & 74.82 &  67.91 & 72.12 \\
        & DKD~\cite{dkd}               
        &  76.70 & 70.35 & 76.45 & 77.07 &  69.71 & 74.06 \\

        &{SimKD~\cite{simkd}}                     
        &  {77.09*} & {67.95*} & {77.18} & {78.39} &  {68.95}  & {73.91} \\
        
        & {DIST~\cite{dist}}               
        &  {-} & {68.66} & {76.34} & {77.35} &  {-} & {74.12} \\ 
        \midrule
        &\textbf{R2KD}   
        & {77.63 (0.06)} & {70.42 (0.03)} & {77.58 (0.08)} & {78.44 (0.03)} & {70.85 (0.11)} & 74.98 \\ 
        &\textbf{{R2KD w/ MixUp}}   
        &  {\underline{77.87} (0.06)} & {\underline{70.70} (0.07)} & {\textbf{78.60} (0.12)} & {\underline{79.06} (0.22)} & {\textbf{71.70} (0.16)} & {\underline{75.58}} \\ 
        &\textbf{R2KD w/ CutMix}   
        &  {\textbf{78.00} (0.16)} & {\textbf{70.87} (0.10)} & {\underline{78.20} (0.14)} & {\textbf{79.44} (0.08)} & {\underline{71.58} (0.14)} &\textbf{75.62} \\ 
        &$\Delta$  
        &  \textbf{+0.86} & \textbf{+0.52} & {\textbf{+1.15}} & {\textbf{+1.05}} & {\textbf{+1.33}} &\textbf{+1.09}\\
        \bottomrule
    \end{tabular}
}
\label{tab:cifar_heterogeneous}
\end{table*}

\subsection{Datasets}

\subsubsection{\textbf{CIFAR-100}~\cite{cifar}} This dataset is widely used for image classification tasks and is publicly available. It contains 100 classes and the samples have an image size of 32$\times$32. The dataset comprises 50000 images in the training set and 10000 images in the test set.

\subsubsection{\textbf{ImageNet}~\cite{imagenet}} This dataset is a massive image classification dataset that contains 1000 classes. The samples are of size 224$\times$224, and the dataset comprises of 1.28 million images in the training set and 5000 images in the test set.

\subsubsection{\textbf{Fine-grained visual recognition (FGVR)}} This dataset present a more difficult challenge. Our experiments are conducted on several such datasets, including Caltech-UCSD Bird (CUB200)~\cite{cub}, MIT Indoor Scene Recognition (MIT67)~\cite{mit67}, Stanford 40 Actions (Stanford40)~\cite{stan40} and Stanford Dogs (Dogs)~\cite{stan_dog}.

\subsubsection{\textbf{TinyImageNet}} This dataset contains small scaled images, which are from ImageNet. Resized images to the same size of CIFAR100 ($32 \times 32$) are used for our experiments.

\subsection{Backbone Networks}
We conducted experiments using popular backbone networks, such as VGG \cite{vgg}, ResNet \cite{resnet}, WRN \cite{wrn}, MobileNet \cite{mobilenet}, and ShuffleNet \cite{shufflenet}, with various teacher-student model combinations including homogeneous and heterogeneous architectures. {For optimization, we train teacher and studnet models using the SGD optimizer for all datasets. For CIFAR-100, TinyImageNet, and FGVR, the weight decay and momentum are set to 5e-4 and 0.9, respectively. For ImageNet, the weight decay and momentum are set to 1e-4 and 0.9, respectively.} It's worth noting that \textbf{all experiments were repeated three times, and the averages were reported.} Implementation details are provided in the supplemental material. The hyper-parameter settings for each datasets are also shown in the supplemental material.

\subsection{Main Results}
We compared our R2KD to various KD methods including logits-based method (KD \cite{kd}, DML~\cite{dml}, TAKD~\cite{takd}, DKD \cite{dkd}, {SimKD~\cite{simkd}, and DIST~\cite{dist}}) and features-based method (FitNet \cite{fitnet}, PKT~\cite{PKT} RKD \cite{RKD}, CRD \cite{CRD}, AT \cite{AT}, VID \cite{VID}, OFD \cite{heo2019comprehensive}, {SP~\cite{sp}}, and ReviewKD~\cite{reviewkd}).

\begin{table*}[t]
\caption{\textbf{Results on classification.} Top-1 and Top-5 accuracy~(\%) on the ImageNet validation. In the row above, ResNet-50 is the teacher and MobileNet-V1 is the student. In the row below, ResNet-34 is the teacher and ResNet-18 is the student.}
\centering
\resizebox{\textwidth}{!}{
\begin{tabular}{ccc|cccc|ccc}
\toprule
\multicolumn{3}{c|}{Distillation} & \multicolumn{4}{c|}{Features}    & \multicolumn{3}{c}{Logits} \\ \hline
   R50-MV1 & Teacher      & Student     & AT\cite{AT}    & OFD\cite{heo2019comprehensive}   & CRD\cite{CRD}   & ReviewKD\cite{revkd} & KD\cite{kd}           & DKD\cite{dkd}  & \textbf{R2KD}         \\ \hline
   top-1      & 76.16        & 68.87       & 69.56 & 71.25 & 71.37 & \underline{72.56}    & 68.58       & 72.05   & 
 \textbf{73.47}       \\
top-5      & 92.86        & 88.76       & 89.33 & 90.34 & 90.41 & 91.00    & 88.98    &   \underline{91.05} &   \textbf{91.61}  

\\ \hline
 R34-R18 & Teacher      & Student     & AT\cite{AT}    & OFD\cite{heo2019comprehensive}   & CRD\cite{CRD}   & ReviewKD\cite{revkd} & KD\cite{kd}     & DKD\cite{dkd}  & \textbf{R2KD}         \\ \hline
top-1      & 73.31        & 69.75       & 70.69 & 70.81 & 71.17 & 71.61    & 70.66 &  \underline{71.70}   & \textbf{72.24}      \\
top-5      & 91.42        & 89.07       & 90.01 & 89.98 & 90.13 & \underline{90.51}    & 89.88 & 90.41  & \textbf{90.65} \\
    \bottomrule
\end{tabular}
}

\label{Table:ImageNet}
\end{table*}

\begin{table*}[]
\caption{\textbf{Results on classification.} Top-1 accuracy (\%) on the FGVR datasets when using teacher and student models with the same and different architectures. The best results are highlighted in \textbf{bold} and the second best \underline{underlined}. {$\Delta$ represents the performance difference between the best result of previous KDs and our best result.}}
\centering
\resizebox{\textwidth}{!}{%
    \begin{tabular}{@{}ccccccccc@{}}
        \toprule
        Dataset & \multicolumn{2}{c}{CUB200} &  \multicolumn{2}{c}{MIT67} &  \multicolumn{2}{c}{Stanford40} & \multicolumn{2}{c}{Dogs}\\ 
        \midrule
        \multirow{2}{*}{Teacher}&  \multicolumn{1}{c}{ResNet34} & \multicolumn{1}{c}{MobileNetV1}  & \multicolumn{1}{c}{ResNet34}  &   \multicolumn{1}{c}{MobileNetV1}    &\multicolumn{1}{c}{ResNet34}  &   \multicolumn{1}{c}{MobileNetV1}    &\multicolumn{1}{c}{ResNet34}  &   \multicolumn{1}{c}{MobileNetV1}    \\ 
         & \multicolumn{1}{c}{61.43} & \multicolumn{1}{c}{67.02}    &\multicolumn{1}{c}{59.55}  &   \multicolumn{1}{c}{61.64}  &  \multicolumn{1}{c}{49.06}  &   \multicolumn{1}{c}{56.06}  &  \multicolumn{1}{c}{69.28}  &   \multicolumn{1}{c}{69.83} \\
        
        \multirow{2}{*}{Student}& \multicolumn{1}{c}{ResNet18} & \multicolumn{1}{c}{ResNet18}  &  \multicolumn{1}{c}{ResNet18}  &   \multicolumn{1}{c}{ResNet18}  &  \multicolumn{1}{c}{ResNet18}  &   \multicolumn{1}{c}{ResNet18}  &  \multicolumn{1}{c}{ResNet18}  &   \multicolumn{1}{c}{ResNet18}    \\ 
        &  \multicolumn{1}{c}{58.14} & \multicolumn{1}{c}{58.14}  &  \multicolumn{1}{c}{57.49}  &   \multicolumn{1}{c}{57.49}  &  \multicolumn{1}{c}{45.94}  &   \multicolumn{1}{c}{45.94}  &  \multicolumn{1}{c}{66.97}  &   \multicolumn{1}{c}{66.97} \\
        \midrule
        FitNet~\cite{fitnet}                      
        & 59.60 &56.00 & 58.28 &57.07 &  46.89 &44.04 &  67.06 &66.25\\
        RKD~\cite{RKD}                
        &  54.80 & 58.80 & 57.63 & 62.14 &  46.68 & 51.12 &  67.23 & 70.49 \\
        CRD~\cite{CRD}                     
        &  60.29 & 64.53 & 59.70 & 63.92 &  49.77 & 54.26 &  68.67 & 70.98 \\
        ReviewKD~\cite{reviewkd}                      
        &  62.13 & 63.09 & 59.68 & 60.76 &  49.95 & 51.77 &  68.96 & 69.22 \\
        \midrule
        KD~\cite{kd}                    
        &  60.92 & 64.74 & 58.78 & 61.87 & 49.42 & 54.07 &  68.28 & 71.82\\
        DKD~\cite{dkd}               
        &  62.17 & 66.45 & 60.00 & 64.35 &  49.84 & 55.80  & 69.04 & 72.53\\ 
        \midrule
        \textbf{R2KD}  
        &  {63.00} & {67.79} & {61.82} & {65.32} &  {50.49} & {56.65} &  {69.75} & {73.45}\\
        
        {\textbf{R2KD w/ MixUp}}   
        & {\underline{63.60}} & {\underline{68.82}} & {\underline{62.29}} & {\underline{66.19}} & {\underline{51.58}} & {\underline{58.00}} & {\underline{70.30}} & {\underline{73.68}}\\ 
        
        \textbf{R2KD w/ CutMix}   
        &  \textbf{63.79} & \textbf{69.65} & \textbf{62.69} & \textbf{66.42} & \textbf{51.90} & \textbf{58.74} & \textbf{70.94} & \textbf{74.06}\\ 
        $\Delta$ 
        &  \textbf{+1.62} & \textbf{+3.20} & \textbf{+2.69} & \textbf{+2.07} &  \textbf{+1.95} & \textbf{+2.94} &  \textbf{+1.90} & \textbf{+1.53}\\
        \bottomrule
    \end{tabular}
}
\label{Table:FGVR}
\end{table*}

\noindent \textbf{CIFAR-100.}
Tables~\ref{tab:cifar_homogeneous} and~\ref{tab:cifar_heterogeneous} presents a summary of the results obtained by using the homogeneous and heterogeneous architecture styles for teacher and student models, respectively. The previous methods were categorized into two types: logits-based models and features-based models, and reported their results from previous studies. The results show that our R2KD are effective in improving performance. In general, the logits-based methods perform worse than the feature-based methods. However, our method, despite being logits-based method, consistently outperforms other features-based methods in all teacher-student pairs. {We also noticed a noticeable increase in performance when combined with the MixUp or CutMix method.} These results are very encouraging, as typical logits-based methods perform poorly when used with CutMix.

\noindent \textbf{ImageNet.} 
The top-1 accuracy of image classification on ImageNet is reported in Table~\ref{Table:ImageNet}.
The results demonstrate that our R2KD achieved significant improvement compared to other distillation methods. Based on the Top-1 accuracy, R2KD obtained performance gains of up to 0.9\% over ReviewKD and up to 1.41\% over DKD.

\noindent \textbf{FGVR.} 
Table~\ref{Table:FGVR} displays the performance evaluation of R2KD on fine-grained visual recognition datasets, which are widely acknowledged to be more challenging. As a result, our framework exhibits state-of-the-art performance on all datasets, for both the same and different teacher-student model pairs. {Additionally, our framework shows even more innovative development when combined with MixUp or CutMix, improving performance by a wide margin.} This will be discussed in more detail in Sec.~\ref{subsec:augmentation}. The results indicate that our framework is able to transfer more abundant knowledge to the student model, even for challenging and augmented samples.

\begin{table*}[]
\caption{\textbf{Effects of data augmentation.} Top-1 accuracy (\%) on the CIFAR-100 dataset with data augmentation. The best results are highlighted in \textbf{bold} and the second best \underline{underlined}. {$\Delta$ represents the performance difference between the best result of previous KDs and our best result.}}
\centering
\resizebox{1.0\textwidth}{!}{%
    \begin{tabular}{@{}ccccccccccc@{}}
    \toprule
        {Teacher}&  \multicolumn{1}{c}{WRN-40-2} & \multicolumn{1}{c}{ResNet56} & \multicolumn{1}{c}{ResNet32x4}  & \multicolumn{1}{c}{VGG13}  &   \multicolumn{1}{c}{VGG13}    &\multicolumn{1}{c}{ResNet50} &\multicolumn{1}{c}{ResNet32x4}  \\         
        {Student}& \multicolumn{1}{c}{WRN-16-2} & \multicolumn{1}{c}{ResNet20} & \multicolumn{1}{c}{ResNet8x4}  & \multicolumn{1}{c}{VGG8}  &   \multicolumn{1}{c}{MobileNetV2}    &\multicolumn{1}{c}{VGG8} &\multicolumn{1}{c}{ShuffleNetV2}  \\         
        \midrule
        KD w/ CutMix~\cite{goodda}                
        & {75.34} & {70.77} & {74.91} & {74.16} & {68.79} & {74.85} & {76.61} \\
                DKD w/ CutMix
        & 75.72 & {71.56} & {76.86} & {75.14} & {70.81} & {75.99} & {78.81} \\
            {CRD w/ CutMix}
        & {75.91} & {71.61} & {75.67} & {74.75} & {70.47} & {75.66} & {76.85} \\
        ReviewKD w/ CutMix
        & {76.00} & 71.14 & 75.91 & 72.72 & 66.88 & 71.24 & 78.78 \\
        \midrule
        KD w/ CutMixPick~\cite{goodda} 
        &  75.59 & 70.99 & 74.78 & 74.43 & 69.49 & 74.95 & 76.90   \\
        CRD w/ CutMixPick~\cite{goodda}
        & 75.96 & 71.41 & 76.11 & 74.65 & 69.95 & 75.35 & 76.93 \\

        \midrule
        {\textbf{R2KD w/ MixUp}}   
        &  {\underline{76.94}} & {\underline{72.60}} & {\textbf{78.07}} & {\underline{75.65}} & {\textbf{71.70}} & {75.80} &  {79.06} \\ 
        
        \textbf{R2KD w/ CutMix}   
        &  \textbf{77.06} & \textbf{72.65} & \underline{77.70} & \textbf{76.40} & \underline{71.58} & \textbf{76.94} & \textbf{79.44} \\
        
        {\textbf{R2KD w/ CutMixPick}}   
        &  {76.72} & {72.19} & {77.52} & {75.54} & {71.28} & {\underline{76.56}} &  {\underline{79.35}} \\ 
        $\Delta$  
        &  \textbf{+1.06} & {\textbf{+1.04}} & {\textbf{+1.21}} & \textbf{+1.26} & {\textbf{+0.89}} & \textbf{+0.95} & 
        {\textbf{+0.63}}\\

        \bottomrule
    \end{tabular}
}
\label{tab:augmenation_cifar}
\end{table*}

\begin{table*}[]
\caption{\textbf{Effects of data augmentation.} Top-1 accuracy (\%) on the TinyImageNet dataset with data augmentation. The best results are highlighted in \textbf{bold} and the second best \underline{underlined}. {$\Delta$ represents the performance difference between the best result of previous KDs and our best result.}}
\centering
\resizebox{1.0\textwidth}{!}{%
    \begin{tabular}{@{}ccccccccccc@{}}
    \toprule
        {Teacher}&  \multicolumn{1}{c}{WRN-40-2} & \multicolumn{1}{c}{ResNet56} & \multicolumn{1}{c}{ResNet32x4}  & \multicolumn{1}{c}{VGG13}  &   \multicolumn{1}{c}{VGG13}    &\multicolumn{1}{c}{ResNet50} &\multicolumn{1}{c}{ResNet32x4}  \\         
        {Student}& \multicolumn{1}{c}{WRN-16-2} & \multicolumn{1}{c}{ResNet20} & \multicolumn{1}{c}{ResNet8x4}  & \multicolumn{1}{c}{VGG8}  &   \multicolumn{1}{c}{MobileNetV2}    &\multicolumn{1}{c}{VGG8} &\multicolumn{1}{c}{ShuffleNetV2}  \\         
        \midrule
        KD w/ CutMix~\cite{goodda}                
        & 59.06 & 53.77 & 56.41 & 62.17 & 60.48 & 61.12 & 67.01 \\
                DKD w/ CutMix
        & 59.92 & 54.01 & 59.23 & 63.12 & {62.73} & 62.84 & {67.97} \\

        {CRD w/ CutMix}
        & {59.42} & {53.68} & {57.40} & {60.36} & {61.17} & {58.80} & {66.68} \\
        
        ReviewKD w/ CutMix
        & 59.96 & {55.04} & 58.01 & 59.92 & 60.30 & 57.69 & 67.66 \\
        \midrule
        KD w/ CutMixPick~\cite{goodda} 
        &  59.22 & 53.66 & 56.82 & 62.32 & 60.53 & 61.40 & 67.08   \\
        CRD w/ CutMixPick~\cite{goodda}
        &  {60.72} & 54.99 & {59.65} & {63.39} & 62.54 & {62.85} & 67.64   \\

        \midrule

        {\textbf{R2KD w/ CutOut}}   
        &  {60.87} & {55.30} & {60.48} & {64.13} & {63.41} & {62.64} & {68.53} \\ 

        {\textbf{R2KD w/ AutoAugment}}   
        &  {60.84} & {54.95} &  {\textbf{60.79}} & {63.90} & {63.36} & {63.30} & {68.85} \\ 
        
        {\textbf{R2KD w/ MixUp}}   
        &  {61.15} & {{55.36}} & {{60.16}} & {{63.39}} & {{63.30}} & {{63.54}} & {{69.08}} \\ 

        \textbf{R2KD w/ CutMix}   
        &  \textbf{61.32} & \underline{55.68} &  {60.56} & \underline{64.69} & \underline{63.52} & \textbf{64.88} & \underline{69.20} \\

        {\textbf{R2KD w/ CutMixPick}}   
        &  {\underline{61.24}} & {\textbf{55.73}} &  {\underline{60.75}} & {\textbf{64.72}} & {\textbf{64.10}} & {\underline{63.58}} & {\textbf{69.23}} \\ 
        
        $\Delta$  
        &  \textbf{+0.60} & {\textbf{+0.69}} &  {\textbf{+1.14}} & {\textbf{+1.33}} & {\textbf{+1.37}} & \textbf{+2.03} & 
        {\textbf{+1.26}}\\

        \bottomrule
    \end{tabular}
}
\label{tab:augmenation_tiny}
\end{table*}

\subsection{Robustness on Augmented Data}
\label{subsec:augmentation}

The data augmentation helps the performance and robustness of the models improve. We utilized {various data augmentation methods such as CutOut~\cite{cutout}, AutoAugment~\cite{autoaug}, MixUp~\cite{mixup}, CutMix~\cite{cutmix}, and CutMixPick~\cite{goodda}.} We used augmented data for additional input alongside the original training set. It is important to note that the number of training samples increases; however, the pre-trained teacher model does not require additional training for the augmented samples. Therefore, to distill the student model successfully, we need to utilize the knowledge of the teacher models that were pre-trained solely on non-augmented samples 

As shown in Table~\ref{tab:augmenation_cifar}, our method, R2KD, outperforms other models with CutMix and CutMixPick by allowing the teacher model to consider not only linear, but also non-linear correlations on the augmented samples. We further boosted the performance by leveraging the network pruning to optimize for learning with data augmentation. The effectiveness of network pruning will be covered in Sec.~\ref{subsec:ablation}. As a result, it is important to properly handle dark knowledge, even when processing inputs in which data augmentation is used.

{In Table~\ref{tab:augmenation_tiny}, we experimented with additional data augmentation methods on TinyImageNet, a notoriously difficult dataset, and found that our model outperformed previous models even when using CutOut, AutoAugment, and MixUp, which are generally known to be less effective than CutMix.} In case ResNet50 and VGG8 are used as the teacher model and student model, respectively, they outperform the latest method, CutMixPick, known for effectively handling augmented data in the field of knowledge distillation by $2.03\%$. This demonstrates that the R2KD effectively processes augmented data, even when dealing with small-scaled challenging images.

\vspace{-0.1cm}

\subsection{Entropy Analysis}
\label{subsec:entropy}

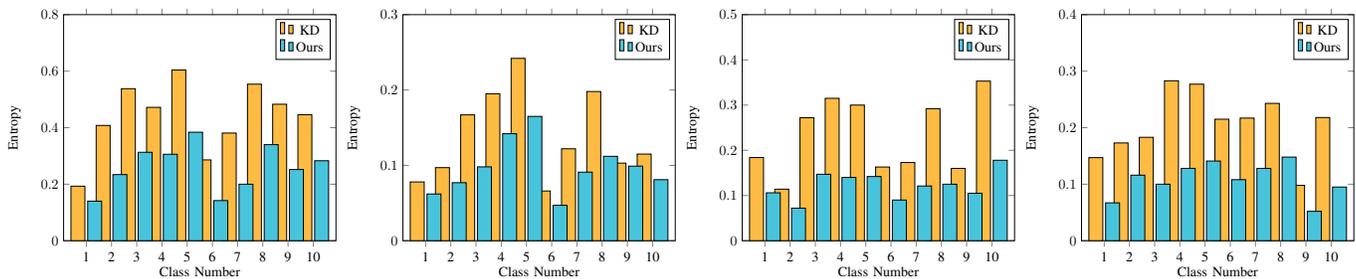
\begin{figure*}[th]
    \centering
    \resizebox{\linewidth}{!}{
\begin{tikzpicture}[scale=0.70]
\begin{axis}[
    ybar,
    xlabel={Class Number},
    ylabel={Entropy},
    ymin=0,
    ymax=0.8,
    symbolic x coords={1,2,3,4,5,6,7,8,9,10},
    xtick=data,
    ]
    \addplot[fill=Dandelion] coordinates {
        (1, 0.193)
        (2, 0.408)
        (3, 0.538)
        (4, 0.472)
        (5, 0.604)
        (6, 0.286)
        (7, 0.381)
        (8, 0.554)
        (9, 0.483)
        (10, 0.446)
    };
    \addplot[fill=SkyBlue] coordinates {
        (1, 0.140)
        (2, 0.234)
        (3, 0.313)
        (4, 0.306)
        (5, 0.384)
        (6, 0.142)
        (7, 0.200)
        (8, 0.340)
        (9, 0.252)
        (10, 0.283)
    };
\legend{KD, Ours}
\end{axis}
\end{tikzpicture}
\begin{tikzpicture}[scale=0.70]
\begin{axis}[
    ybar,
    xlabel={Class Number},
    ylabel={Entropy},
    ymin=0,
    ymax=0.3,
    symbolic x coords={1,2,3,4,5,6,7,8,9,10},
    xtick=data,
    ]
    \addplot[fill=Dandelion] coordinates {
        (1, 0.078)
        (2, 0.097)
        (3, 0.167)
        (4, 0.195)
        (5, 0.242)
        (6, 0.066)
        (7, 0.122)
        (8, 0.198)
        (9, 0.103)
        (10, 0.115)
    };
    \addplot[fill=SkyBlue] coordinates {
        (1, 0.062)
        (2, 0.077)
        (3, 0.098)
        (4, 0.142)
        (5, 0.165)
        (6, 0.047)
        (7, 0.091)
        (8, 0.112)
        (9, 0.099)
        (10, 0.081)
    };
\legend{KD, Ours}
\end{axis}
\end{tikzpicture}
\begin{tikzpicture}[scale=0.70]
\begin{axis}[
    ybar,
    xlabel={Class Number},
    ylabel={Entropy},
    ymin=0,
    ymax=0.5,
    symbolic x coords={1,2,3,4,5,6,7,8,9,10},
    xtick=data,
    ]
    \addplot[fill=Dandelion] coordinates {
        (1, 0.184)
        (2, 0.114)
        (3, 0.272)
        (4, 0.315)
        (5, 0.300)
        (6, 0.163)
        (7, 0.173)
        (8, 0.292)
        (9, 0.160)
        (10, 0.353)
    };
    \addplot[fill=SkyBlue] coordinates {
        (1, 0.106)
        (2, 0.072)
        (3, 0.147)
        (4, 0.140)
        (5, 0.142)
        (6, 0.090)
        (7, 0.121)
        (8, 0.125)
        (9, 0.105)
        (10, 0.178)
    };
\legend{KD, Ours}
\end{axis}
\end{tikzpicture}
\begin{tikzpicture}[scale=0.70]
\begin{axis}[
    ybar,
    xlabel={Class Number},
    ylabel={Entropy},
    ymin=0,
    ymax=0.4,
    symbolic x coords={1,2,3,4,5,6,7,8,9,10},
    xtick=data,
    ]
    \addplot[fill=Dandelion] coordinates {
        (1, 0.147)
        (2, 0.173)
        (3, 0.183)
        (4, 0.283)
        (5, 0.277)
        (6, 0.215)
        (7, 0.217)
        (8, 0.243)
        (9, 0.098)
        (10, 0.218)
    };
    \addplot[fill=SkyBlue] coordinates {
        (1, 0.067)
        (2, 0.116)
        (3, 0.100)
        (4, 0.128)
        (5, 0.141)
        (6, 0.108)
        (7, 0.128)
        (8, 0.148)
        (9, 0.052)
        (10, 0.095)
    };
\legend{KD, Ours}
\end{axis}
\end{tikzpicture}
}
   
    \caption{\textbf{Comparison of entropy.} Entropy for several classes that have high entropy from teacher model. Left: ResNet32x4-ResNet8x4, Left-Middle: ResNet32x4-ShuffleNetV2, Right-Middle: VGG13-VGG8, Right: ResNet50-MobileNetV2.}
    \label{entropy}
\end{figure*}

\begin{figure*}[th]
\centering
\includegraphics[width=0.90\textwidth]{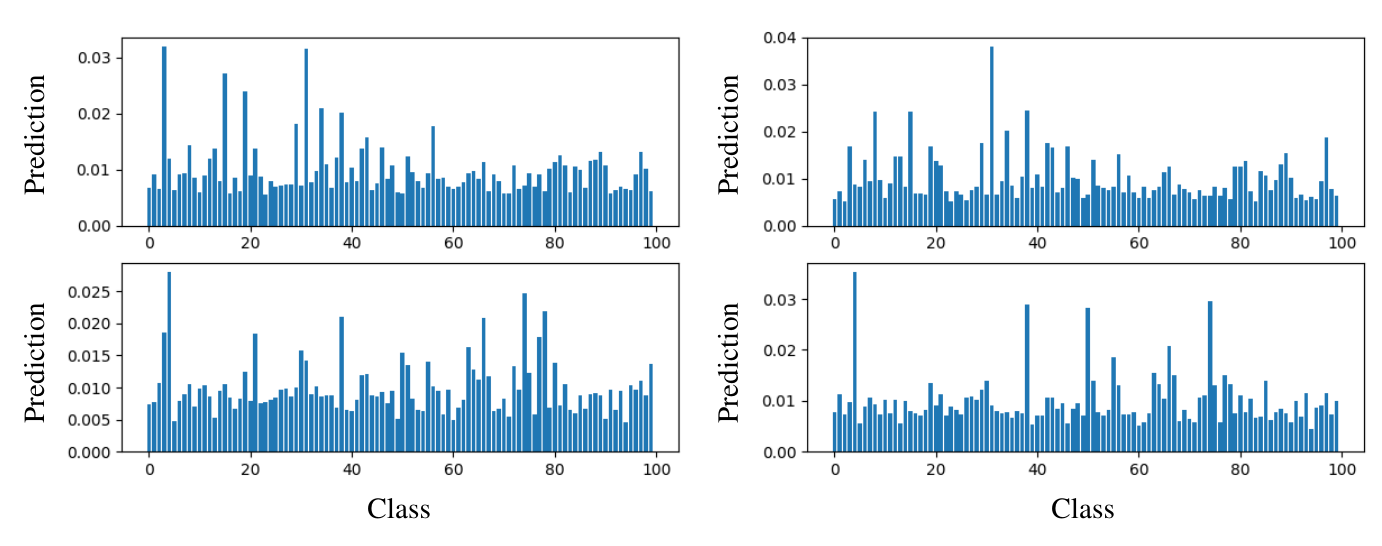}
\caption{\textbf{Comparison of entropy.} Prediction distributions for the samples with high entropy extracted from the testset of CIFAR-100. The teacher is ResNet32x4 and student is ResNet8x4. }
\label{fig:dark}
\end{figure*}

We investigated the impact of R2KD on the distribution of dark knowledge. To accomplish this, we employed the teacher model to classify CIFAR-100 image samples based on high and low entropy. Samples with {low} entropy exhibit overly confident target predictions, leading to reduced distillation performance due to insufficient dark knowledge (like yellow box in Fig.~\ref{fig:correlation}). Conversely, samples with {high} entropy possess excessive dark knowledge, which also adversely affects distillation performance due to a lack of target prediction information (like red box in Fig.~\ref{fig:correlation}). Therefore, maintaining an appropriate balance of dark knowledge is crucial for achieving optimal distillation~\cite{ATS}.

To demonstrate that our method effectively transfers optimal information from the teacher model, we analyzed the entropy of predictions obtained by the student models, selecting only the samples with high entropy from the teacher model, i.e., those with insufficient target knowledge. {To identify these high entropy samples, we calculated the entropy for the entire dataset and categorized the samples with the top 10\% entropy as high entropy samples.} Fig.~\ref{entropy} displays the average entropy about the 10 classes, with our method yielding lower entropy compared to traditional KD. This suggests that our model carries more reliable target information, resulting in improved student accuracy. This finding is also consistent with Fig.~\ref{fig:fkl} (a).

Additionally, Fig.~\ref{fig:dark} illustrates the prediction distribution obtained using DKD and R2KD for high entropy samples identified by the teacher. This demonstrates that R2KD enhances the target predictions for these samples, leading to a clearer identification of the correct label. Therefore, the correlation distance, including value- and rank-based correlation, has the ability to optimize the performance of the student model by providing the appropriate distillation distribution for each sample.

\begin{table*}[]
\caption{\textbf{Ablation studies.} The experiments are conducted on CIFAR-100, ImageNet, and MIT67. The evaluation metric is Top-1 accuracy $(\%)$. $\mathcal{L}_\text{Value}$: Value-based distance loss. $\mathcal{L}_\text{Rank}$: Rank-based distance loss. $\boldsymbol{p}^\text{Pr}$: Pruned teacher model. In each dataset, the final row represents the performance difference between the results without any of our methods and the results with all of our methods.}
		\centering
		\resizebox{0.8\textwidth}{!}{%
		\begin{tabular}{cccccc}
			\toprule
			Datasets & $\mathcal{L}_\text{Value}$ & $\mathcal{L}_\text{Rank}$ & Pruning $\left(\boldsymbol{p}^\text{Pr}\right)$ & ResNet56 / ResNet20 & ResNet32x4 / ResNet8x4 \\ \midrule
			 \multirow{6}{*}{CIFAR-100} & $\square$ & $\square$ & $\square$ & 70.66 & 73.33  \\
            & $\square$ & $\square$ & $\blacksquare$ & {71.38} & {73.95} \\ 
            & $\blacksquare$ & $\square$ & $\square$ & 71.92 & 76.51 \\
   		 & $\blacksquare$ & $\blacksquare$ & $\square$ & 72.26 & 76.81  \\
			 & $\blacksquare$ & $\blacksquare$ & $\blacksquare$ & \textbf{72.42} & \textbf{77.01}  \\
      & & & & \textbf{+1.76} & \textbf{+3.68} \\

    \midrule
    			 &  &  &  & ResNet50 / MobileNetV1 & ResNet34 / ResNet18 \\ \midrule
			 \multirow{4}{*}{ImageNet} & $\square$ & $\square$ & $\square$ & 68.58 & 70.66  \\
       		 & $\square$ & $\square$ & $\blacksquare$ & 72.67 & 71.98  \\
   		 & $\blacksquare$ & $\blacksquare$ & $\blacksquare$ & \textbf{73.47} & \textbf{72.24}  \\
      & & & & \textbf{+4.89} & \textbf{+1.58} \\
    \midrule
    			 &  &  &  & MobileNetV1 / ResNet18 & ResNet34 / ResNet18 \\ \midrule
			 \multirow{4}{*}{MIT67}& $\square$ & $\square$ & $\square$ & 61.87 & 58.78  \\
              & $\blacksquare$ & $\blacksquare$ & $\square$ & 65.37 & 61.07  \\
   		 & $\blacksquare$ & $\blacksquare$ & $\blacksquare$ & \textbf{66.19} & \textbf{61.82}  \\
      & & & & \textbf{+4.32} & \textbf{+3.04} \\
    \bottomrule
		\end{tabular}
		
		}
		\label{tab:ablation}
\end{table*}

\subsection{Ablation Study}
\label{subsec:ablation}
To demonstrate the effectiveness of each proposed method, we performed ablation studies on the CIFAR100, ImageNet and MIT67 datasets. For the results of CIFAR 100, Table~\ref{tab:ablation} shows that performance improvement of $3.68~\%$ in case ResNet32x4 and ResNet8x4 are used for the teacher and the student model, respectively. Also, in case of ImageNet, the second row shows the results when adding only network pruning on the ImageNet dataset. Compared to the baseline model, we observed a performance gain of $1.32~\%$ in accuracy for ResNet34-ResNet18 and a performance gain of $4.09~\%$ in accuracy for ResNet50-MobileNetV1. The reason for the improved performance is that clear samples maintain the high-confidence score of target prediction, while ambiguous samples reduce the confidence score. A detailed network pruning analysis is described in supplemental material. Furthermore, the thrid row shows the performance considering value-based and rank-based correlation distance. For ResNet34-ResNet18, we observed a performance increase of $1.58~\%$ in Top-1 accuracy over KD, and for ResNet50-MobileNetV1, we observed a performance improvement of $4.89~\%$ in Top-1 accuracy. The reason for this performance improvement is that we can consider both linear and non-linear relationships, which cannot be accounted for by using traditional KL-divergence. This trend can also be shown on the MIT67 dataset, the second row shows the results of adding value-based and rank-based correlation distance. Compared to KD, we observed a performance gain of $2.29~\%$ in accuracy for ResNet34-ResNet18 and a performance gain of $3.5~\%$ in accuracy for MobileNetV1-ResNet18. The third row about MIT67 shows the performance with network pruning. For ResNet34-ResNet18, we observed a performance gain of $3.04~\%$ in Top-1 accuracy over KD, and for MobileNetV1-ResNet18, we observed a performance gain of $4.32~\%$ in Top-1 accuracy.

\subsection{Visualizations}

\begin{figure}[]
	\centering
	{\includegraphics[width=1.0\linewidth]{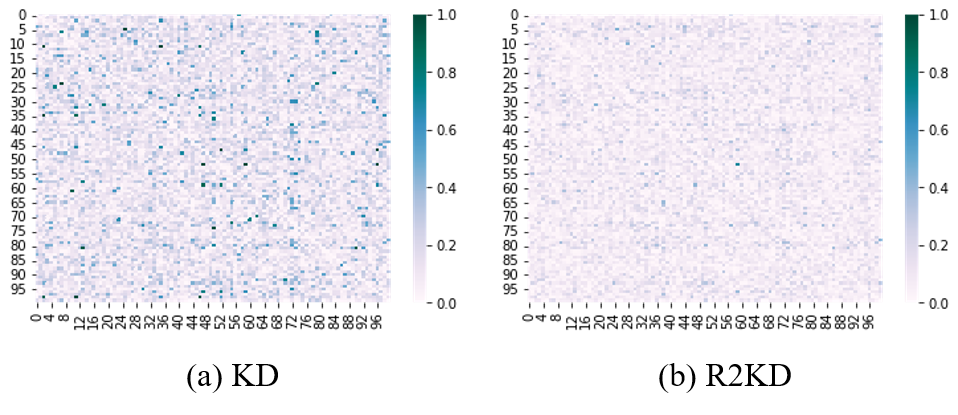}}
	\caption{\textbf{Disparities in correlation matrices} between the logits of the student and teacher. Our R2KD show smaller disparities than KD.}
	\label{fig:difference}
\end{figure}

\begin{figure}[]
	\centering
	{\includegraphics[width=0.9\linewidth]{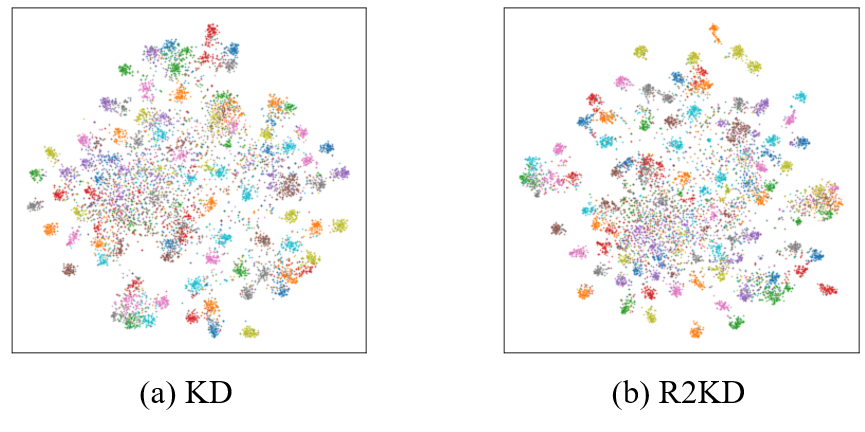}}
	\caption{\textbf{tSNE} of features from KD and R2KD}
	\label{fig:tsne}
\end{figure}

We provide visualizations from two viewpoints, using ResNet32x4 as the teacher and ResNet8x4 as the student on CIFAR-100. First, Fig.~\ref{fig:difference} display visual representations of the disparities in correlation matrices between the logits of the student and teacher. In contrast to KD, R2KD encourages the student to produce logits that are more similar to those of the teacher, thereby achieving superior distillation performance. Additionally, Fig.~\ref{fig:tsne} shows the t-SNE results, which indicate that the representations produced by R2KD are more distinguishable compared to KD, confirming that R2KD enhances the discriminability of deep features. 

{To complement the visual insights provided by the t-SNE plots, we also calculated homogeneity, completeness, and V-score to offer a more precise and quantitative evaluation of the clustering ability. Homogeneity measures how well each cluster contains only data points from a single class, while completeness assesses how well all data points of a single class are assigned to the same cluster. The V-score, as the harmonic mean of these two metrics, provides a comprehensive measure of clustering quality.}

{For the ResNet32x4 and ResNet8x4 model pair, the KD method achieved scores of 0.955, 0.958, and 0.956 for homogeneity, completeness, and V-score, respectively. In contrast, our method achieved scores of 0.9619, 0.9615, and 0.9616 for these metrics. Similarly, for the WRN-40-2 and SV1 model pair, the KD method obtained scores of 0.9597, 0.9601, and 0.9599, while our method achieved scores of 0.9610, 0.9635, and 0.9622. These results demonstrate the superiority of our method over the KD method.}

\subsection{{Comparison of Multiple Teacher and Pruned Teacher}}

{To compare the use of multiple teacher models with a simple ensemble method and the use of pruned teachers, we measured performance by adding differently initialized teacher models and employing a weighted sum between them. We also evaluated performance by increasing the number of pruned teachers. As indicated in Table~\ref{Table:ensemble_teacher}, the performance was comparable when using multiple teacher models and when using pruned teachers. However, while the ensemble method requires additional time to train new teacher models, pruned teachers are obtained without a retraining process by merely removing the less significant weights from the pre-trained teacher's weights, making it a more efficient method than the ensemble technique. Furthermore, it was verified that utilizing two pruned teacher models leads to further performance improvements compared to two additional teacher models. This suggests that using network pruning methods can effectively reduce the risk of incorrect predictions from the original teacher, resulting in better performance enhancements.}


\subsection{{Complexities and Performance of Networks}}

{Table~\ref{tab:complexity_student} and Table~\ref{tab:complexity_teacher} show the complexity of the student and teacher models, respectively. We use the pre-trained teacher model to obtain a pruned model, then take the weighted sum of the predictions from both models for logit-based distillation. As shown in Table~\ref{tab:complexity_teacher}, there is only a minor performance difference between the pruned model and the original model because we do not retrain the pruned model; we simply remove the less important weights from the original model.
Furthermore, it should be noted that the original and pruned models have the same number of parameters and the same number of FLOPs because we applied unstructured pruning. Unstructured pruning sets some of the weights in the model to zero; it does not reduce the number of parameters in the layers themselves. Additionally, it does not impact FLOPs because the computations are still performed even if the weights are zeroed out. However, in terms of inference time, the pruned teacher model with a pruning rate of 0.1 exhibited approximately 0.3 times the inference time compared to the original teacher model. This indicates that, despite having the same FLOPs, the pruned teacher model demonstrates higher time efficiency because sparse computations take less time.}

\begin{table}[]
	\caption{{Accuracy with multi-teachers on the CIFAR-100 dataset. Additional teachers are obtained by training with different initialized weights without applying network pruning. For pruned teacher models, we use pruning rates of $\{0.01, 0.05, 0.1\}$.}}
	\centering
 
	\resizebox{0.8\columnwidth}{!}{
		\begin{tabular}{c|c}
			\toprule
			\textbf{Teacher} & \textbf{Accuracy} \\
			\midrule
            R2KD w/o pruned teacher model & 76.81 \\
            	\midrule
            + 1 teacher & 77.02 \\
            + 2 teacher & 77.06 \\
            	
		+ 1 pruned teacher & 77.01 \\
            + 2 pruned teacher & 77.22 \\
            + 3 pruned teacher & 77.29 \\
           			
			\bottomrule
		\end{tabular}
	}

	\label{Table:ensemble_teacher}
\end{table}

\begin{table*}[t]
\caption{{Complexities and performance of student networks on CIFAR-100 dataset.}}
\centering
\resizebox{0.7\textwidth}{!}{%
    \begin{tabular}{@{}cccccccccc@{}}
    \toprule
        {Teacher}&  \multicolumn{1}{c}{WRN-16-2} & \multicolumn{1}{c}{ResNet20} & \multicolumn{1}{c}{ResNet8x4}  & \multicolumn{1}{c}{VGG8}  &   \multicolumn{1}{c}{MobileNetV2}    &\multicolumn{1}{c}{ShuffleNetV2}\\  
        
        \midrule
        
        \# Params
        & {0.70M} & {0.28M} & {1.23M} & {3.97M} & {0.81M} & {1.36M} \\
        \midrule
        FLOPs
        & {102.33M} & {41.62M} & {178.58M} & {96.73M} & {7.37M} & {46.76M} \\
        \midrule
        Accuracy
        & {73.26} & {69.06} & {72.50} & {70.36} & {64.60} & {71.82} \\

        \bottomrule
    \end{tabular}
}
\label{tab:complexity_student}
\end{table*}

\begin{table*}[t]
\caption{{Complexities and performance of teacher networks on CIFAR-100 dataset.}}
\centering
\resizebox{0.6\textwidth}{!}{%
    \begin{tabular}{@{}ccccccccc@{}}
    \toprule
        {Teacher}&  \multicolumn{1}{c}{WRN-40-2} & \multicolumn{1}{c}{ResNet56} & \multicolumn{1}{c}{ResNet50}  & \multicolumn{1}{c}{ResNet32x4}  &   \multicolumn{1}{c}{VGG13}  \\  
        
        \midrule
        
        \# Params
        & {2.26M} & {0.86M} & {23.71M} & {7.43M} & {9.46M}  \\
        \midrule
        FLOPs
        & {330.66M} & {127.93M} & {1311.78M} & {1088.22M} & {285.99M}  \\
        \midrule
        Original Teacher
        & {75.61} & {72.34} & {79.34} & {79.42} & {74.64} \\
        Pruned Teacher
        & {75.50} & {72.34} & {79.31} & {79.42} & {74.56} \\

        \bottomrule
    \end{tabular}
}
\label{tab:complexity_teacher}
\end{table*}

\section{Conclusions}
This paper identified a negative issue with the use of KL divergence in knowledge distillation, which can lead to the transfer of inappropriate information based on the teacher's entropy, subsequently resulting in reduced student performance. To address this challenge, we projected the distributions of both the teacher and student models into a vector space and introduced correlation distance into the loss function, thereby encouraging the alignment of the student vector with the direction of the teacher vector. Our proposed method, Robustness-Reinforced Knowledge Distillation (R2KD), consistently demonstrated performance improvements, even when dealing with challenging and heavily augmented datasets. To further enhance the robustness of the student model, we incorporated network pruning into the teacher model. We extensively validated our method on various datasets, including CIFAR-100, FGVR, TinyImageNet, and ImageNet, demonstrating its superior accuracy and robustness compared to other existing KD methods. We hope that our R2KD approach will serve as a foundational advancement for the integration of data augmentation techniques into the knowledge distillation process, thereby further improving the efficacy of model compression and knowledge transfer in practical applications.


\medskip

\newpage    
{
\small
\bibliography{egbib}
\bibliographystyle{plain}
}

\vfill

\end{document}